\documentclass[letterpaper]{article}
\usepackage{aaai}
\usepackage{times}
\usepackage{helvet}
\usepackage{courier}

\usepackage{graphicx}
\usepackage{longtable}
\usepackage{url}
\usepackage{listings}
\usepackage{xcolor}
\usepackage{amsmath}
\usepackage{natbib}

\usepackage{todonotes}

\definecolor{mygreen}{rgb}{0,0.6,0}
\definecolor{mygray}{rgb}{0.5,0.5,0.5}
\definecolor{mymauve}{rgb}{0.58,0,0.82}
\definecolor{myred}{rgb}{0.9,0.2,0.2}

\newcommand{\specReq}[1]{\ensuremath{\mathtt{^{#1}}}}

\lstdefinelanguage{pddl}
{
  sensitive=false,    
  morecomment=[l]{;}, 
  alsoletter={:,-},   
  morekeywords={
    define,domain,problem,not,and,or,when,imply,forall,exists,either,
    :domain,:extends,:requirements,:types,:objects,:constants,
    :constraints,:ordered-substasks,:subtasks,:tasks,
    :predicates,:action,:durative-action,:duration,:method,:durative-method,
    :htn,:parameters,:precondition,:condition,:effect,:functions,
    :fluents,:primary-effect,:side-effect,:init,:goal,assign
    :strips,:adl,:equality,:task,:typing,:conditional-effects,:metric,
    :negative-preconditions,:disjunctive-preconditions,
    :existential-preconditions,:universal-preconditions,:ordered-subtasks,:ordering
  },
  keywords=[2]{object,at,start,over,all,end,always,at-most-once,sometime-before,sometime,sometime-after,hold-during,hold-between,hold-after,hold-before,minimize,maximize,total-time}, 
  keywords=[3]{calib_direction,image_direction,instrument,satellite,mode}, 
  keywords=[4]{calibrate,turn_approx,turn_precise,take_image,turn_to,activate_instrument,point_to,take_video,method_stereo,do_observation_stereo, do_observation,decrease_overall_quality,nethod_observe}, 
  keywords=[5]{observable,calibrated,pointing,supports,power_on,power_avail,on_board,calib_target,have_image,
  image-quality,calib-time,turn-time} 
}

\begin{document}

%
 \title{HDDL 2.1: Towards Defining an HTN Formalism with Time}

 \author{D. Pellier\footnotemark[1], H. Fiorino\footnotemark[1], M. Grand\footnotemark[1], A. Albore\footnotemark[2], R. Bailon-Ruiz\footnotemark[2]\\
  \footnotemark[1]University Grenoble Alpes, LIG, Grenoble, France\\
  \{damien.pellier,humbert.fiorino,maxence.grand\}@imag.fr\\
  \footnotemark[2]ONERA, DTIS, Toulouse, France\\
  \{alexandre.albore,rafael.bailon\_ruiz\}@onera.fr\\
 }

\maketitle

\begin{abstract}
\begin{quote}

Real world applications of planning, like in industry and robotics, require modelling rich and diverse scenarios. Their resolution usually requires coordinated and concurrent action executions. In several cases, such planning problems are naturally decomposed in a hierarchical way and expressed by a Hierarchical Task Network (HTN) formalism.
The PDDL language used to specify planning domains has evolved to cover the different planning paradigms. However, formulating real and complex scenarios where numerical and temporal constraints concur in defining a solution is still a challenge.
Our proposition aims at filling the gap between existing planning languages and operational needs. To do so, we propose to extend HDDL taking inspiration from  PDDL 2.1 and ANML to express temporal and numerical expressions. This paper opens discussions on the semantics and the syntax needed to extend HDDL, and illustrate these needs with the modelling of an Earth Observing Satellite planning problem.


\end{quote}
\end{abstract}

\subsection{Introduction}

The Hierarchical Task Network (HTN) formalism \citep{Erol94} is used to express a wide variety of planning problems in many real-world applications, e.g., in task allocation for robot fleets \citep{Milot21}, video games \citep{Menif14} or industrial contexts such as software deployment \citep{Georgievski17}. Real world applications of planning, like in industry and robotics, require modelling rich and diverse scenarios. In several cases, such  planning problems are naturally decomposed in a hierarchical way, and they contain both numerical and temporal constraints defining the sub-tasks decomposition, and agents synchronisation on collaborative tasks. In fact, concurrency between the actions, and agents coordination in HTN problems are needed to find solutions for nontrivial tasks in complex scenarios.

Over the last years, much progress has been made in the field of hierarchical planning \citep{bercher19}. Novel systems based on the traditional, search-based techniques have been introduced~\citep{Bit-Monnot:16,Ramoul17,Shivashankar17,Bercher17,Holler19,Holler20,Holler21}, but also new techniques like the translation to STRIPS/ADL~\citep{Alford09,Alford16,behnke2022}, or revisited approaches like the translation to propositional logic~\citep{Behnke2018totSAT,Behnke2019orderchaos,Schreiber2019SAT,Schreiber21,behnke2021}.

Despite these advances, not all these systems use the same formalism to represent hierarchical task decomposition. This makes difficult to compare the different approaches and to promote HTN planning techniques, even if some formalism are more adopted than others. For these reasons, an extension of the PDDL (Planning Domain Description Language)~\citep{mcdermott98},  used in the International Planning Competitions (IPC), has been proposed. This extension \citep{holler:19}, called HDDL (Hierarchical Planning Domain Description Language), is based on PDDL 2.1 \citep{Fox03} and is the result of several discussions within the planning community \citep{Holler19b} to fill the need of a standard language for the first Hierarchical Planning track of IPC 2020. In this first version, it was decided that, considering the efforts to develop the language and related tools, none of the temporal and numerical features of PDDL would be included. The objective of this paper is to initiate and organise the discussion on the temporal extension of HDDL, to propose some starting ideas and tools.

Our motivation is grounded on compelling needs to devise applications involving autonomous systems, i.e. agents, which must be coordinated to fulfil hierarchical tasks containing both numerical and temporal constraints. As a matter of fact, action concurrency and agent coordination are needed in HTN problems to find solutions for nontrivial tasks in complex scenarios, like collaborative robotic missions \citep{weser2010htn}. Starting from a real satellite application example, we propose two extensions for HDDL including elements of PDDL 2.1 and ANML to express temporal and numerical expressions. This is intended to initiate discussions within HTN community on establishing a standard -- HDDL 2.1~-- aimed at filling the gaps between existing hierarchical-temporal planning approaches. To that end, we make this preliminary extension of HDDL an open source project with a publicly available repository.

The rest of the paper is organised as follows. Section 1 and section 2 propose two extensions of HDDL 1.0 syntax to deal with time, numeric and constraints in method declaration. Section 3 illustrates HDDL 2.1 with a real world example of an Earth Observing Satellite (EOS) mission. Section 4 presents the related works, and Section 5 discusses the main features of the proposed language. The full syntax of our HDDL dialect is presented in the appendix.



\section{Durative methods}

%
%

To deal with time in HDDL, we propose to add the concept of durative method. Like methods in HDDL 1.0, durative methods have parameters, the abstract task they decompose, and the resulting task network, and, like durative actions, they can have duration constraints and a condition that has to hold to apply the decomposition. Note that, unlike durative actions, duration constraints may be not mandatory. By default the duration of a method is the sum of all the durations of the primitive tasks that compose it. However, it can be interesting to specify temporal constraints on the execution of the decomposition such as, for example, whatever decomposition is chosen, the duration of the primitive tasks must not exceed a given value, or that a subtask must last less than another subtask. We propose to allow these durative constraints only if the requirement {\tt :durative-inequalities} is set. This is the same requirement as for PDDL.

\begin{lstlisting}[firstnumber=last, escapechar=~]
<durative-method-def> ::=~\label{l:durative-methods}~
    :durative-method <name>
    :parameters (<typed list (variable)>)
    :task (<task-symbol> <term>*)
    [:duration <method-duration-constraint>]~\specReq{:duration-inequalities}~
    [:condition <da-gd>]~\specReq{:method-preconditions}~~\label{l:mcond}~
    <tasknetwork-def>~\label{l:msubtasks2}~)
\end{lstlisting}

\subsection{Method Duration Constraints}

We propose to add optional constraints to restrict the allowed value for the duration of the method or the sub-tasks that compose it. Therefore, it could be necessary to be able to refer explicitly to the duration of a subtask and not only to the duration of the method with the variable {\tt ?duration}. We propose the following syntax for the method duration constraints based on PDDL syntax.

\begin{lstlisting}[firstnumber=last, escapechar=~]
<method-duration-constraint> ::=~\specReq{:duration-inequalities}~ (and~\linebreak~ <method-simple-duration-constraint>+)
<method-duration-constraint> ::= ()
<method-duration-constraint> ::= ~\mbox{<simple-method-duration-constraint>}~
<simple-method-duration-constraint>::= ~\linebreak~(<binary-comp> <duration> <td-value>)
<simple-method-duration-constraint>::= ~\linebreak~(at <time-specifier> ~\mbox{<simple-method-duration-constraint>}~)
<duration> :: = ?duration
<duration> :: = <task-duration>
<td-value> ::= <d-value>
<td-value> ::= <task-duration>
<task-duration> :: = (duration <task-id>)
\end{lstlisting}

\subsection{Method Condition}

We propose to used the same syntax as PDDL temporal precondition of the durative method (see \verb|<da-gd>| rule in appendix).

\subsection{Method Ordering Constraints}

To deal with time, it is necessary to generalise the ordering constraints defined in the task network of the method. Task network declaration stays unchanged from HDDL 1.0. We just propose to declare the ordering constraints via the task ids decorated with time specifiers. Time specifiers allow to specify as in a point algebra if the ordering constraints are about the start time or the end time of the task. We propose to use the same time specifiers as in PDDL: ~{\tt start} to define the start time of a task, and {\tt end} to define the end time of a task. Note that we propose to extend also the list of operators available to express ordering constraints between tasks. It has been extended to define any constraints on the start or end of a sub-task of a method, and allows to express all Allen's intervals \citep{allen1981interval} between these subtasks.

\begin{lstlisting}[firstnumber=last, escapechar=~]
<ordering-defs> ::= () | <ordering-def> ~\linebreak~| (and <ordering-def>+)
<ordering-def> ::= ~\linebreak~(<d-task> <task-id> <task-id>)
<ordering-def> ::=~\specReq{:durative-action} \linebreak~(<d-task> <time-task-id> <time-task-id>)
<d-task> ::= <
<d-task> ::=~\specReq{:durative-action}~>=
<d-task> ::=~\specReq{:durative-action}~<=
<d-task> ::=~\specReq{:durative-action}~>
<d-task> ::=~\specReq{:durative-action}~=
<time-task-id> ::= (<time-specifier> ~\mbox{<task-id>)}~
<time-specifier> ::= start
<time-specifier> ::= end
\end{lstlisting}

\section{HTN Method Constraints and Semantics}

HDDL 1.0 only accepts equality constraints or inequality on method parameters in the task network. We propose to add constraints based on the syntax of the constraints used on state trajectory constraints defined in PDDL 3.0 \citep{gerevini:05}, and that were first introduced by \cite{Erol94}. These constraints allow to express logical constraints on the decomposition of the method in addition to the ordering constraints. In particular, it is possible to show that the preconditions of the durative methods can be rewritten with these constraints without losing the generality of the form of these constraints. We propose to use a new requirements for these constraints: \verb+:method-constraints+.

\begin{lstlisting}[firstnumber=last, escapechar=~]
<constraint-defs> ::= () | <constraint-def> | (and <constraint-def>+)
<constraint-def> ::= (not (= <term> <term>))~\linebreak~| (= <term> <term>)
<constraint-def> ::=~\specReq{:method-constraints}~ ~\linebreak~(hold-before <task-id> <gd>)
<constraint-def> ::=~\specReq{:method-constraints}~ ~\linebreak~(hold-after <task-id> <gd>)
<constraint-def> ::=~\specReq{:method-constraints}~ ~\linebreak~(hold-between <task-id> <task-id> <gd>)
<constraint-def> ::=~\specReq{:method-constraints}~ ~\linebreak~(hold-during <task-id> <task-id> <gd>)
<constraint-def> ::=~\specReq{:method-constraints}~ ~\linebreak~(at end <effect>)
<constraint-def> ::=~\specReq{:method-constraints}~ ~\linebreak~(at start <gd>)
<constraint-def> ::=~\specReq{:method-constraints}~ ~\linebreak~(always <gd>)
<constraint-def> ::=~\specReq{:method-constraints}~ ~\linebreak~(at-most-once <gd>)
<constraint-def> ::=~\specReq{:method-constraints}~ ~\linebreak~(sometime <gd>)
<constraint-def> ::=~\specReq{:method-constraints}~ ~\linebreak~(sometime-before  <task-id> <gd>)
<constraint-def> ::=~\specReq{:method-constraints}~ ~\linebreak~(sometime-after <task-id> <gd>)
\end{lstlisting}

The semantics of these constraints can be defined as in \citep{gerevini:05}. Given a task network $w$ (i.e. a tuple $(T, \prec)$, where $T$ is a set of tasks and $\prec \subseteq T \times T$ is a strict partial order on the tasks), and a state $s_0$, that is the state prior to the execution of $w$, and $\pi$ a decomposition of $N$ into atomic subtasks; $\pi$ generates the finite temporal trajectory:
$$
\langle (s_0, t_0), (s_1, t_1), \ldots, (s_n, t_n) \rangle
$$

This trajectory is valid if it satisfies all the constraints of $tn$. Let then $\phi$ and $\psi$  be two goal description formulae over the predicates of the planning problem. 
We note $[t_s, t_e]T$ the interval of execution of a task $T$. Given these intervals, where $t_s$ is the start time and $t_e$ is the end time, a series of times ordering constraints are defined in an interpretation such as:

 Constraint {\tt hold-before} expresses the properties that must be satisfied in the state before the execution of a specific subtask. It generalises the notion of preconditions
used in classical planning to task networks ($\phi$ is a goal description and $\models$ is the satisfiability relation as defined in \citep{gerevini:05}).
 \begin{multline*}
\langle (s_0, t_0), (s_1, t_1), \ldots, (s_n, t_n) \rangle \models \\
     (\text{\tt hold-before} \ [t_s, t_e]T \ \phi) \  \text{iff} \ s_{t_s} \models \phi
\end{multline*}

 Constraint {\tt hold-after} expresses the properties of the world that must be satisfied after the execution of a specific task.
 \begin{multline*}
\langle (s_0, t_0), (s_1, t_1), \ldots, (s_n, t_n) \rangle \models \\
     (\text{\tt hold-after} \ [t_s, t_e]T \ \phi) \ \text{iff}  \ s_{t_e} \models \phi
\end{multline*}

 Constraints {\tt at start} and {\tt at end} are generalisations of {\tt hold-before} and {\tt hold-after} constraints. They express that a specified expression must be satisfied before or after every subtask of a task network. Both constraints should be added to be compatible with PDDL version 3.1 and to simplify the expression of constraints between subtasks. These two constraints can easily be rewritten in {\tt hold-before} and {\tt hold-after} constraints.
\begin{multline*}
\langle (s_0, t_0), (s_1, t_1), \ldots, (s_n, t_n) \rangle \models \\
    (\text{\tt at end } \phi) \;\; \text{iff} \; s_n \models \phi
\\
\shoveleft\langle (s_0, t_0), (s_1, t_1), \ldots, (s_n, t_n) \rangle \models\\
     (\text{\tt at start} \ \phi) \;\; \text{iff} \; s_0 \models \phi \qquad\qquad\quad
\end{multline*}
Constraint {\tt hold-between} indicates that some properties must always remain true between the end of the execution of a task, and until the begin of the execution of another task. This type of constraint makes it possible to express the notion of causal link used in Plan Space Planning.
\begin{multline*}
 \langle (s_0, t_0), \ldots, (s_i, t_i), \ldots, (s_n, t_n) \rangle \models \\
    (\text{\tt hold-between} \ [t^1_s, t^1_e]T_1 \ [t^2_s, t^2_e]T_2 \ \phi) \\
\text{iff} \ \forall t_i : t^1_e \leq t_i \leq t^2_s \ \text{such as} \ s_{i} \models \phi
\end{multline*}
Constraint {\tt hold-during} is quite similar. However, it does not cover the same interval. For {\tt hold-during}, some properties must always remain true during the execution of a task until the end of the execution of another task. This constraint is not strictly necessary, but is necessary to be compatible with PDDL version 3.1. It can be rewritten in terms of constraints {\tt hold-before}, {\tt hold-between} and {\tt hold-after}. It allows to specify temporal ``envelops''.
\begin{multline*}
 \langle (s_0, t_0), \ldots, (s_i, t_i), \ldots, (s_n, t_n) \rangle \models \\
  (\text{\tt hold-during } [t^1_s, t^1_e]T_1 \ [t^2_s, t^2_e]T_2 \ \phi) \\
  \text{iff} \ \forall t_i : t^1_s \leq t_i \leq t^2_e \ \text{such as} \ s_{i} \models \phi
\end{multline*}
Constraint {\tt always} expresses that some properties must be  satisfied whatever the decomposition of the task network is.
\begin{multline*}
\langle (s_0, t_0), \ldots, (s_i, t_i), \ldots, (s_n, t_n) \rangle \models \\
 (\text{\tt always } \phi) \ \text{iff} \ \forall t_i : t_0 \leq t_i \leq t_n \ s_i \models \phi
\end{multline*}
Constraint {\tt sometime} say that some properties must be verified sometime in the trajectory produced by the decomposition of the task network.
\begin{multline*}
\langle (s_0, t_0), \ldots, (s_i, t_i), \ldots, (s_n, t_n) \rangle \models  (\text{\tt sometime} \: \phi) \\
\text{iff} \quad  \exists t_i: t_0 \leq t_i \leq t_n \: \text{such as} \: s_i \models \phi
\end{multline*}
Constraint {\tt sometime-before} says that some properties must be verified sometime before the begin of the execution of a specific subtask of the task network.
\begin{multline*}
\langle (s_0, t_0), \ldots, (s_i, t_i), \ldots, (s_n, t_n) \rangle \models   \\
\left(\text{\tt sometime-before} \; [t_s, t_e]T \: \phi\right) \qquad\\
\qquad\qquad\qquad\text{iff} \; \exists t_i: t_0 \leq t_i \leq t_n \; \text{such as} \; s_i \models \phi \: \ \text{and} \; t_i \leq t_s
\end{multline*}
Constraint {\tt sometime-after} says that some properties must be verified sometime after the end of the execution of a specific subtask of the task network.
\begin{multline*}
\langle (s_0, t_0), \ldots, (s_i, t_i), \ldots, (s_n, t_n) \rangle \models \\
(\text{\tt sometime-after} \ [t_s, t_e]T \ \phi) \qquad \\
\shoveright{ \qquad \qquad \qquad \text{iff} \ \exists t_i: t_0 \leq t_i \leq t_n \ \text{such as} \ s_i \models \phi \ \text{and} \ t_i \geq t_e}
\end{multline*}
Constraint {\tt at-most-once} indicates that some properties must be satisfied at most once in the trajectory produced by the decomposition of the task network.
\begin{multline*}
\langle (s_0, t_0), \ldots, (s_n, t_n) \rangle \models (\text{\tt at-most-once} \ \phi) \\
 \text{iff} \ \forall t_i: t_0 \leq t_i \leq t_n \ \text{if} \ s_i \models \phi \ \text{then} \\
 \exists t_j: t_j \geq t_i \ \text{such as} \ \forall t_k: t_i \leq t_k \leq t_j \ \text{such as} \\
\ s_k \models \phi \text{ and} \ \forall t_k : t_k > t_j \ \text{such as} \ s_k \models \neg\phi
\end{multline*}

Such timed constraints are an important aspect of the semantics we discuss here. Other aspects are left out for space reasons, but are open questions that deserve to be discussed for a  timed HDDL standard, namely the treatment of empty methods and actions, the way time-constraints and durations are dealt with by method's decompositions.


%
%
%


\section{A satellite example}

Our reflection on HTN planning with time are based on a commercial optical Earth Observing Satellite (EOS) study case, which delimits our requirements for this new language, namely \emph{durative actions} and \emph{durative methods}, \emph{method preconditions}, \emph{numeric preconditions}, and \emph{timed ordering constraints}.

We present thereafter this real use case of an EOS with the mission of observing areas on the surface of the planet and embedding the capacity of producing and executing autonomous plans. This example is shaped on the International Planning Competition hierarchical satellite domain~\citep{schattenberg2020}, and is intended to illustrate the temporal features, as well as the orderings and the hierarchical constraints of the proposed language extension.

New generations of commercial EOSs guarantee better performances by embedding a certain degree of autonomy on-board~\citep{pralet2019}. An automated planner is expected to produce new observation mission plans whenever environment changes or new requests are sent from ground control. Because observation tasks depend on the environment ---which is not known in advance--- this is a planning problem and not just a scheduling matter~\citep{rodriguez-moreno2004}.

\subsection{Description of the domain}

The EOS mission goal is to fulfil a set of \emph{Acquisition Requests} of images of the Earth surface. Requests can be single images, which can be obtained by powering on sensors, and pointing at the sites of interest. The different modalities of fulfilling those requests require the extension of the satellite domain to a temporal model.

Considering the structure of the requests that must be completed,
the action specifications have a \emph{duration} to reflect the warming up of certain instruments and the time to modify the attitude of the satellite. Figure \ref{pddl_calibrate} shows the calibrate action having a duration specific to the used instrument and defined in the problem initial state.
%
The action has conditions that, if set \texttt{at start} are equivalent to the Classical planning PDDL preconditions. In a similar way, effects set to happen \texttt{at end} are applied at the end of the action duration, similarly to classical planning effects.
\begin{figure}[!h]
\begin{lstlisting}[language=pddl,mathescape]
(:durative-action calibrate
	:parameters (?c_s - satellite ?c_i - instrument ?c_d - calib_direction)
	:duration (= ?duration (calib-time ?c_i))
	:condition 	(and
			(at start (on_board ?c_i ?c_s))
			(at start (calib_target ?c_i ?c_d))
			(at start (pointing ?c_s ?c_d))
			(at start (power_on ?c_i))	)
	:effect
		(and  (at end (calibrated ?c_i))) )
\end{lstlisting}
\caption{The calibrate action has a duration depending on the instrument calibration time\label{pddl_calibrate}}
\end{figure}

Calibration times are defined in the initial state, e.g.\\
{\small\lstinline[language={pddl},basicstyle=\ttfamily]|(= (calib-time instrument0) 20)|}.

The satellite is agile, that is, it can rotate forwards and backwards in addition to left and right with respect to its track. This capability is used to point towards \textit{sites} that represent the direction of an acquisition request. However this is only possible during a short period of time $[t^{start}, t^{end}]$ in which the location is visible. The observable timespan of each {\tt site} of interest is specified in the initial situation with timed initial literals like in PDDL 2.2~ \citep{hoffmann2005deterministic}, e.g.

 \begin{figure}[!h]
 \begin{lstlisting}[language=pddl]
 (:objects
 	site - image_direction
 )
 (:init
 	(at 500 (observable site))
 	(at 1000 (not (observable site)))
     ...
 ) \end{lstlisting}
 \caption{Timed initial literals for site observability\label{pddl_initplan_dom}}
 \end{figure}

 The \texttt{take-image} preconditions depend then on the site visibility (Figure~\ref{pddl_take-image}).
 This action has conditions that should hold at start like in Fig.~\ref{pddl_calibrate}, but also \emph{throughout the action execution}, noted with \texttt{over all}; hence it should be a durative-action, with a duration (here set to a constant value).

\begin{figure}[!h]
\begin{lstlisting}[language=pddl]
(:durative-action take_image
	:parameters (?ti_s - satellite
	    ?ti_d - image_direction
	    ?ti_i - instrument ?ti_m - mode)
	:duration (= ?duration 1.00000)
	:precondition (and
		  (over all (observable ?ti_d))
		  (at start (calibrated ?ti_i))
		  (at start (pointing ?ti_s ?ti_d))
		  (over all (on_board ?ti_i ?ti_s))
		  (over all (power_on ?ti_i))
		  (at start (supports ?ti_i ?ti_m))	)
	:effect	(and
		(at end (have_image ?ti_d ?ti_m))
		(at end (increase
		   (image-quality ?ti_s) 2))) )
\end{lstlisting}
\caption{\texttt{take\_image} conditions depend on site observability.\label{pddl_take-image}}
\end{figure}
Pointing to sites is performed by the \texttt{turn\_to} task. \texttt{turn\_to} can be decomposed in two ways, one of which is less precise (the overall quality of the images will be reduced) but quicker to execute. These two \emph{durative-methods} differ mainly in the duration constraints, reflecting the different pointing times. This choice, only possible on-board, corrects over-confident plans regarding transition times that otherwise would have caused the acquisition to fail. The planning problem can be designed to optimise image quality, but also accept lower quality images when time is short. On the modelling side, the distinction between the two methods is done
on the basis of the duration interval.
\begin{figure}[!h]
\begin{lstlisting}[language=pddl]
(:durative-method turn_precise
	:parameters (?tp_s - satellite
	    ?tp_d_prev - image_direction
	    ?tp_d - image_direction)
	:task (turn_to ?tp_s ?tp_d ?tp_d_prev)
	:duration (>= ?duration (* (/ (turn-time
	    (t_d_new t_d_prev)) 4) 5) )
	:ordered-subtasks (and
		(point_to ?tp_s ?tp_d ?tp_d_prev)
		(take_image ?tp_s ?tp_d ?tp_i ?tp_m) )
	:constraints (and
		(not (= ?tp_d ?tp_d_prev))) )

(:durative-method turn_approx
	:parameters (?ta_s - satellite
	    ?ta_d_prev - image_direction
	    ?ta_d - image_direction)
	:task (turn_to ?ta_s ?ta_d ?ta_d_prev)
	:duration (and
		  (<= ?duration (* (/ (turn-time
		      (t_d_new t_d_prev)) 4) 5))
		  (>= ?duration (/ (turn-time
		      (t_d_new t_d_prev)) 2) )
	:ordered-subtasks (and
		(point_to ?ta_s ?ta_d ?ta_d_prev)
		(take_image ?ta_s ?ta_d ?ta_i ?ta_m)
		(decrease_overall_quality ?ti_s) )
	:constraints (and
		(not (= ?ta_d ?ta_d_prev))) )
\end{lstlisting}
\caption{\texttt{turn\_precise} and \texttt{turn\_approx} methods use duration inequalities to reflect the duration time, which is bounded by the duration of the \emph{durative-action} \texttt{point\_to}. The duration of \texttt{point\_to} is bounded by the relative position of the two sites to observe.\label{pddl_durative_m}}
\end{figure}

Methods (pre)conditions can be a way, like method durations, to filter, or to anticipate, the decomposition of the subtasks.
A simple way to use them, is to set them to the precondition of the first subtask, when the ordering of the tasks is known.
In a more expressive way, in the following example (Figure~\ref{pddl_durative_m_con}),
the durative-method constraints fix an order between the subtasks task0 and task1 when the duration of the latter is lower than the calibration time of the instrument (line~\ref{l:mdur}) by imposing that the effect of instrument activation holds before task1 (line~\ref{l:mhb}). Similarly, we allow to constrain subtask durations, which is a powerful feature aimed at providing a finer control on the method decomposition, e.g. by using
\mbox{\small\lstinline[language={pddl},basicstyle=\ttfamily]|(< (duration task1) (duration task0))|} in the temporal constraints of Figure~\ref{pddl_durative_m_con} example.

\begin{figure}[th]
\begin{lstlisting}[language=pddl, basicstyle=\fontsize{8.5}{10}\selectfont\ttfamily, escapechar=~]
	(:durative-method method_observe
		:parameters (?ta_d_prev - image_direction
		   ?ta_s - satellite
		   ?ta_d - image_direction
		   ?ta_i - instrument ?m - mode)
		:task (do_observation ?ta_d ?m)
		:duration
		 (< (duration task1) (calib-time ?ta_i))~\label{l:mdur}~
		:subtasks (and
		 (task0 (activate_instrument ?ta_s ?ta_i))
		 (task1 (turn_to ?ta_s ?ta_d ?ta_d_prev))
		 (task2 (take_image ?ta_s ?ta_d ?ta_i ?m)))
		:ordering (and
			(< task0 task2)
			(< task1 task2)	)
		:constraints (and
			(not (= ?ta_d ?ta_d_prev))
			(hold-before task1 (power_on ?ta_i)))
	)~\label{l:mhb}~
\end{lstlisting}
\caption{\texttt{method\_observe} uses a duration inequality and a constraint over literals to constrain an ordering. Similarly, a decomposition can be constrained at method-level.\label{pddl_durative_m_con}}
\end{figure}

\section{Related work}

Durative actions with intermediate effects have received some criticisms in the past~\citep{smith03}, in particular the fact that they fail in guaranteeing that preconditions/conditions hold over specified intervals, and that effects can take place at arbitrary time points within the action. Even if it is true that forcing all effects to take place at the start and
end points of actions, as in PDDL2.1 and as we intend to do, imposes constraints on how activity can be modelled, it does not represent any constraint on expressiveness \citep{fox2004}. While PDDL2.1 is not the most convenient language for durative actions with intermediate effects, HDDL supports them with ease as they can be naturally divided into sub-actions.
Here, the extension to durative methods is simple, considering a method duration is the max of the duration of its decompositions.

In 2008, a first attempt to standardise a hierarchical and temporal language was proposed. This language, called ANML (Action Notation Modeling Language) \citep{smith08}, is an effort to provide a high level, convenient, and succinct alternative to existing languages. ANML is based on the notions of action and state, uses a variable/value representation, supports temporal constraints, and provides simple and convenient idioms to express the most common forms of action conditions and effects. However, few  planners adopt the ANML formalism as input, e.g., FAPE \citep{dvorak2014} or EUROPA \citep{Barreiro12}. The main drawback of ANML is that it is not as widespread as the PDDL family of languages, which are more familiar to the planning community. Therefore, extending HDDL ---that is based on PDDL--- to deal with temporal and numerical problems seems to be a better approach to standardise and structure the planning community.

Several temporal hierarchical and numeric planners have already been presented in the literature. One of the first is TimeLine \citep{yaman2002}. This planner is an extension of the well known and widely used HTN planner SHOP2 \citep{nau2003}. It supports actions with durative effects. Task decompositions are described as methods that have preconditions, subtasks, and time constraints (with inequalities) between subtasks. Another temporal extension of SHOP2 was proposed by \cite{goldman2006}. This provides a way to encode PDDL durative actions in a HTN formalism compatible with the SHOP2 planner. The same year, \cite{castillo2006} published a new hierarchical temporal planner called SIADEX. It exploits partial order relations to obtain parallel plans, using PDDL durative actions based on Simple Temporal Networks. The formalism used by SIADEX, although based on PDDL, is specific to it and has not been taken over by other planners.

\section{Discussion}

With the assistance of an EOS planning problem inspired from an IPC benchmark and grounded on a real application, we have illustrated the main features that HDDL 2.1 aimed at describing hierarchical temporal domains should integrate.
The generation of EOS mission plans is highly combinatorial as multiple activity types --- instrument operation, payload data download and maintenance --- are to be planned under strict resource and time constraints.
In order to reflect the temporal constraints present in satellite operations, we have adopted a problem formulation that includes elements of temporal planning, besides the usual HDDL 1.0 formulation for expressing hierarchical planning problems~\citep{holler:19}.

This effort aims at ferrying HTN planning towards domains close to real world applications, where the temporal elements like concurrent actions, coordination, and a hierarchical distribution of tasks are prominent.

The lack of a unified language featuring both temporal and numerical constraints in the continuity of the works done on PDDL is, in our view, an obstacle and a need to address. The automated planning community has produced in the past several approaches for modelling complex planning problems including temporal aspects and a hierarchical decomposition of tasks, but the variety of language solutions is also a hindrance for having common tools and solvers.

ANML  \citep{smith08} has an expressivity close to what we seek here.
In ANML effects that happen at time intervals during an action duration can be specified. This  can also be represented in HDDL~2.1 by using constraints semantics and dividing actions with intermediate effects into separate durative actions.

FAPE proposes benchmarks where ordering constraints are delayed.\footnote{E.g. \url{https://github.com/laas/fape/blob/master/planning/domains/rovers_ipc5-hier/rovers_ipc5-hier.dom.anml#L180}.} For instance in ANML it is possible to specify that an action must happen at least some amount of time after the end of a previous action. With the proposed syntax and semantics, such expressivity can be reached by using auxiliary tasks that decompose in a durative primitive task (of the desired duration) with no effects.

Time sampling represents another open question for this extension of HDDL with time.
Basically, two approaches exist. Sampling can be either constant ---when time is divided into regular-spaced discrete steps--- or with variable time steps instantiated when effects and preconditions are applied. The latter can benefit from the Simple Temporal Problem formalism to model the temporal aspects of the plan and to include timed initial effects, and Interval Temporal Logics can be used to define truth of formulas relative to time intervals, rather than time points \citep{BRESOLIN2014269}.

Joined with the suggested HDDL 2.1 language grammar, we propose a parser and an illustrative benchmark part of the PDDL4J librairie \citep{Pellier18} available online\footnote{\url{https://github.com/pellierd/pddl4j}}. Future works include, as first step, an open discussion on the semantics and the syntax. To that end, we publish the current preliminary version of HDDL 2.1 as an open source project publicly available on a repository: {\it link will be provided in the final version of the paper}. More work lies ahead to provide the planning community with the tools it needs to deal with hierarchical temporal planning problems. As for instance developing a plan validator for HDDL 2.1.

To be fully compatible with PDDL 3.0 features, the language HDDL 2.1 needs to include axioms and preferences, besides the associated parsing, and validating tools.
Benchmarks in HDDL 2.1 are also something we plan to provide as challenges for future solvers and the International Planning Competition (IPC). To provide a rich variety of HTN temporal instances, we plan to decompose hierarchically the already existing benchmarks of the temporal IPC track or to translate existing ANML domains from the FAPE distribution~\citep{dvorak2014}.  Such collection of planning domains can be the kernel of a future IPC for HDDL 2.1 solvers.



\bibliographystyle{aaai}
\bibliography{ref}

\appendix
\section{Appendix: Full Proposal of Syntax for HDDL~2.1}

This proposed syntax for HDDL2.1.alpha takes its roots in HDDL~\citep{holler:19}. From that initial definitions, we here describe its proposed extension.

\subsection{Domain Description}

The domain definition has been extended to durative actions (line~\ref{l:durative-actions}) and durative methods (line~\ref{l:durative-methods}).

\begin{lstlisting}[escapechar=~]
<domain> ::= (define (domain <name>)
    [<require-def>]
    [<types-def>]~\specReq{:typing}~
    [<predicates-def>]
    [<functions-def>]~\specReq{:fluents}~
    [<constants-def>]
    [<task-defs>]
    <structure-def>*)
\end{lstlisting}

%
%
%

\noindent The definition of the basic domain elements is nearly unchanged. Function declaration are now possible.

\begin{lstlisting}[firstnumber=last, escapechar=~]
<require-def> ::= ~\linebreak~(:requirements <require-key>+)
<require-key> ::= ~\textit{see below}~
<types-def> ::= (:types ~\mbox{<typed-list(<primitive-type>))}~
<predicates-def> ::= ~\linebreak~(:predicates <atomic-formula-skeleton>+)
<function-def> ::= ~\linebreak~(:<function-typed-list (atomic-function-skeleton)>)
<constants-def> ::= ~\linebreak~(:constants <typed-list (constant)>)
<atomic-formula-skeleton> ::= ~\linebreak~(<predicate> <typed-list(variable)>)
<atomic-function-skeleton> ::= ~\linebreak~(<function>~\linebreak~ <typed-list (variable)>)
<typed-list (x)> ::= x+ - <type> ~\linebreak~[<typed list (x)>]~\label{l:typedlist}~
<type> ::= (either <primitive-type>+)
<type> ::= <primitive-type>
<function-typed-list (x)> ::= x+ - ~\linebreak~<function-type> <function-typed-list(x)>
<function typed list (x)> ::=
<function-type> ::=~\specReq{:numeric-fluents}~number
<function-type> ::=~\specReq{:typing} \specReq{:object-fluents}~<type>
<predicate> ::= <name>
<function> ::= <name>
<constant> ::= <name>
<variable> ::= ?<name>
<primitive-type> ::= <name>
<primitive-type> ::= object
\end{lstlisting}

%
%
\noindent Abstract tasks are defined similarly to HDDL~1.0.   
\begin{lstlisting}[firstnumber=last, escapechar=~]
<tasks-def> ::= (:task <task-def>)
<task-def> ::= <task-symbol> ~\linebreak~:parameters (<typed list (variable)>)~\label{l:compTask}~
<task-symbol> ::= <name>
\end{lstlisting}

\noindent The structure of a temporal HDDL domain is composed of durative and non-durative methods and actions. Durative and non-durative actions are defined as in PDDL and non-durative methods as in HDDL. 

\begin{lstlisting}[firstnumber=last, escapechar=~]
<structure-def> ::= <action-def>
<structure-def> ::=~\specReq{:durative-actions}~ ~\mbox{<durative-action-def>}~
<structure-def> ::= <method-def>
<structure-def> ::=~\specReq{:durative-methods}~ ~\mbox{<durative-method-def>}~
\end{lstlisting}

%
%
\noindent  The method declaration remains unchanged from HDDL~1.0.

\begin{lstlisting}[firstnumber=last, escapechar=~]
<method-def> ::= (:method <name>~\label{l:methods}~
    :parameters (<typed list (variable)>)~\label{l:mparams}~
    :task (<task-symbol> <term>*)~\label{l:mabstask}~
    [:precondition <gd>]~\specReq{:method-preconditions}~~\label{l:mprec}~
    <tasknetwork-def>~\label{l:msubtasks}~)
\end{lstlisting}

%
%
\noindent Durative methods are added.

\begin{lstlisting}[firstnumber=last, escapechar=~]
<durative-method-def> ::=~\label{l:durative-methods}~
    :durative-method <name>
    :parameters (<typed list (variable)>)
    :task (<task-symbol> <term>*)
    [:duration <method-duration-constraint>]~\specReq{:duration-inequalities}~
    [:condition <da-gd>]~\specReq{:method-preconditions}~~\label{l:mcond}~
    <tasknetwork-def>~\label{l:msubtasks2}~)
\end{lstlisting}

%
%
\noindent The task network declaration remains unchanged from HDDL~1.0.
\begin{lstlisting}[firstnumber=last, escapechar=~]
<tasknetwork-def> ::=
    [~\textbf{:}~[~\textbf{ordered-}~][~\textbf{sub}~]~\textbf{tasks}~
        <subtask-defs>]~\label{l:tnsubtasks}~
    [~\textbf{:order}~[~\textbf{ing}~] <ordering-defs>]~\label{l:tnordering}~
    [:constraints <constraint-defs>]~\label{l:tnconstraints}~
\end{lstlisting}

%
%
\noindent The subtask definition can contain one or more subtasks. 
\begin{lstlisting}[firstnumber=last, escapechar=~]
<subtask-defs> ::= () | <subtask-def> ~\linebreak~| (and <subtask-def>+)
<subtask-def> ::= (<task-symbol> <term>*) ~\linebreak~| (<subtask> (<task-symbol> <term>*))
<subtask> ::= <name>
\end{lstlisting}

%
%
\noindent The ordering constraints declaration are extended to deal time feature;

\begin{lstlisting}[firstnumber=last, escapechar=~]
<ordering-defs> ::= () | <ordering-def> ~\linebreak~| (and <ordering-def>+)
<ordering-def> ::= ~\linebreak~(<d-task> <task-id> <task-id>)
<ordering-def> ::=~\specReq{:durative-action} \linebreak~(<d-task> <time-task-id> <time-task-id>)
<d-task> ::= <
<d-task> ::=~\specReq{:durative-action}~>=
<d-task> ::=~\specReq{:durative-action}~<=
<d-task> ::=~\specReq{:durative-action}~>
<d-task> ::=~\specReq{:durative-action}~=
<time-task-id> ::= (<time-specifier> ~\mbox{<task-id>)}~ 
<time-specifier> ::= start
<time-specifier> ::= end
\end{lstlisting}

%
%
Methods constraints are extended based on the syntax of the constraints used on state trajectory constraints defined in PDDL 3.0.

\begin{lstlisting}[firstnumber=last, escapechar=~]
<constraint-defs> ::= () | <constraint-def> | (and <constraint-def>+)
<constraint-def> ::= (not (= <term> <term>))~\linebreak~| (= <term> <term>)
<constraint-def> ::=~\specReq{:method-constraints}~ ~\linebreak~(hold-before <task-id> <gd>)
<constraint-def> ::=~\specReq{:method-constraints}~ ~\linebreak~(hold-after <task-id> <gd>)
<constraint-def> ::=~\specReq{:method-constraints}~ ~\linebreak~(hold-between <task-id> <task-id> <gd>)
<constraint-def> ::=~\specReq{:method-constraints}~ ~\linebreak~(hold-during <task-id> <task-id> <gd>)
<constraint-def> ::=~\specReq{:method-constraints}~ ~\linebreak~(at end <effect>)
<constraint-def> ::=~\specReq{:method-constraints}~ ~\linebreak~(at start <gd>)
<constraint-def> ::=~\specReq{:method-constraints}~ ~\linebreak~(always <gd>)
<constraint-def> ::=~\specReq{:method-constraints}~ ~\linebreak~(at-most-once <gd>)
<constraint-def> ::=~\specReq{:method-constraints}~ ~\linebreak~(sometime <gd>)
<constraint-def> ::=~\specReq{:method-constraints}~ ~\linebreak~(sometime-before  <task-id> <gd>)
<constraint-def> ::=~\specReq{:method-constraints}~ ~\linebreak~(sometime-after <task-id> <gd>)
\end{lstlisting}

%
%
\indent The original action definitions defined in PDDL remains unchanged. 

\begin{lstlisting}[firstnumber=last, escapechar=~]
<action-def> ::= (:action <task-def>~\label{l:action}~
    [:precondition <gd>]
    [:effect <effect>])
\end{lstlisting}

%
%

\noindent Durative action are now possible based on the same syntax as PDDL~2.1.
\begin{lstlisting}[firstnumber=last, escapechar=~]
<durative-action-def> ::= ~\linebreak~(:durative-action <task-def>~\label{l:durative-actions}~
      :duration <duration-constraint> 
      [:condition <da-gd)>]
      [:effect <da-effect>])
\end{lstlisting}

%
%
As in PDDL, duration constraints with the \verb|:duration-inequalities| requirement allow to express duration inequalities. The definition of the duration constraints for durative actions does not change.

\begin{lstlisting}[firstnumber=last, escapechar=~]
<action-duration-constraint> ::=~\specReq{:duration-inequalities}~(and ~\linebreak~ <simple-action-duration-constraint>+)
<action-duration-constraint> ::= ()
<action-duration-constraint> ::= ~\mbox{<simple-action-duration-constraint>}~
<action-simple-duration-constraint>::= ~\linebreak~(<d-op> ?duration <d-value>)
<action-simple-duration-constraint>::= ~\linebreak~(at <time-specifier> ~\mbox{<simple-duration-constraint>}~)
<d-op> :: <=
<d-op> :: >=
<d-op> :: =
<d-value> ::= <number>
<d-value> ::=~\specReq{:numeric-fluents}~<f-exp>
<f-exp> ::=~\specReq{:numeric-fluents}~<number>
<f-exp> ::=~\specReq{:numeric-fluents}~(<binary-op> <f-exp> <f-exp>)
<f-exp> ::=~\specReq{:numeric-fluents}~(<multi-op> <f-exp> <f-exp>+)
<f-exp> ::=~\specReq{:numeric-fluents}~(- <f-exp>)
<f-exp> ::=~\specReq{:numeric-fluents}~<f-head>
<f-head> ::= (<function-symbol> <term>*)
<f-head> ::= <function-symbol>
<binary-op> ::= <multi-op>
<binary-op> ::= -
<binary-op> ::= /
<multi-op> ::= *
<multi-op> ::= +
<binary-comp> ::= >
<binary-comp> ::= <
<binary-comp> ::= =
<binary-comp> ::= >=
<binary-comp> ::= <=
\end{lstlisting}

%
%
\noindent We proposed the following syntax for the the duration constraints of between method sub-tasks.

\begin{lstlisting}[firstnumber=last, escapechar=~]
<method-duration-constraint> ::=~\specReq{:duration-inequalities}~ (and~\linebreak~ <method-simple-duration-constraint>+)
<method-duration-constraint> ::= ()
<method-duration-constraint> ::= ~\mbox{<simple-method-duration-constraint>}~
<simple-method-duration-constraint>::= ~\linebreak~(<binary-comp> <duration> <td-value>)
<simple-method-duration-constraint>::= ~\linebreak~(at <time-specifier> ~\mbox{<simple-method-duration-constraint>}~)
<duration> :: = ?duration 
<duration> :: = <task-duration>
<td-value> ::= <d-value> 
<td-value> ::= <task-duration>
<task-duration> :: = (duration <task-id>)
\end{lstlisting}

%
%
\noindent The goal description of HDDL~1.0 is proposed to be extened to deal with numeric fluent.
\begin{lstlisting}[firstnumber=last, escapechar=~]
<gd> ::= ()
<gd> ::= <literal (term)>
<gd> ::= (and <gd>*)
<gd> ::=~\specReq{:disjunctive-preconditions}~ (or <gd>*)
<gd> ::=~\specReq{:negative-preconditions}~ (not <gd>)
<gd> ::=~\specReq{:disjunctive-preconditions \ :negative-preconditions}~ ~\mbox{(imply <gd> <gd>)}~
<gd> ::=~\specReq{:existential-preconditions}~ ~\linebreak~(exists (<typed list (variable)>*) <gd>)
<gd> ::=~\specReq{:universal-preconditions}~ ~\linebreak~(forall (<typed list (variable)>*) <gd>)
<gd> ::= (= <term> <term>)
<gd> ::=~\specReq{:numeric-fluents} <f-comp>
<f-comp> ::= (<binary-comp> <f-exp> <f-exp>)
\end{lstlisting}

\begin{lstlisting}[firstnumber=last, escapechar=~]
<literal (t)> ::= <atomic formula(t)>
<literal (t)> ::= (not <atomic formula(t)>)
<atomic formula(t)> ::= (<predicate> t*)
\end{lstlisting}

\begin{lstlisting}[firstnumber=last, escapechar=~]
<term> ::= <name>
<term> ::= <variable>
\end{lstlisting}

%
%

\noindent Now, we propose to use timed goal description defined in PDDL~2.1 for durative action precondition and durative method condition.

\begin{lstlisting}[firstnumber=last, escapechar=~]
<da-gd> ::= <timed-gd>
<da-gd> ::= (and <da-gd>*)
<da-gd> ::=~\specReq{:universal-preconditions}~ (forall ~\linebreak~ <typed-list (variable)>) <da-gd>)
<timed-gd> ::= (at <time-specifier> <gd>)
<timed-gd> ::= (over <interval> <gd>)
<interval> ::= all
\end{lstlisting}

%
%
\noindent Symmetrically, we allow the used the used of temporal effect as in PDDL~2.1.

\begin{lstlisting}[firstnumber=last, escapechar=~]
<effect> ::= ()
<effect> ::= (and <c-effect>*)
<effect> ::= <c-effect>
<c-effect> ::=~\specReq{:conditional-effects}~ ~\linebreak~(forall (<variable>*) <effect>)
<c-effect> ::=~\specReq{:conditional-effects}~ ~\linebreak~(when <gd> <cond-effect>)
<c-effect> ::= <p-effect>
<p-effect> ::= (not <atomic formula(term)>)
<p-effect> ::= <atomic formula(term)>
<cond-effect> ::= (and <p-effect>*)
<cond-effect> ::= <p-effect>
\end{lstlisting}

%
%

\noindent The temporal and numeric definitions of effects are as follows:

\begin{lstlisting}[firstnumber=last, escapechar=~]
<da-effect> ::= (and <da-effect>*)
<da-effect> ::= <timed-effect>
<da-effect> ::=~\specReq{:conditional-effects}~ (forall ~\linebreak~(<typed list (variable)>) <da-effect>)
<da-effect> ::=~\specReq{:conditional-effects}~ ~\linebreak~(when <da-gd> <timed-effect>)
<timed-effect> ::= ~\linebreak~(at <time-specifier> <cond-effect>) 
<timed-effect> ::=~\specReq{:numeric-fluents}~ ~\linebreak~(at <time-specifier> <f-assign-da>)
<timed-effect> ::=~\specReq{:continuous-effects + :numeric-fluents}~ ~\linebreak~(<assign-op-t> <f-head> <f-exp-t>)
<f-assign-da> ::= (<assign-op> <f-head> <f-exp-da>)
<f-exp-da> ::= (<binary-op> <f-exp-da> <f-exp-da>)
<f-exp-da> ::= (<multi-op> <f-exp-da> <f-exp-da>+)
<f-exp-da> ::= (- <f-exp-da>)
<f-exp-da> ::=~\specReq{:duration-inequalities}~ ?duration
<f-exp-da> ::= <f-exp>
<assign-op-t> ::= increase
<assign-op-t> ::= decrease
<f-exp-t> ::= (* <f-exp> #t)
<f-exp-t> ::= (* #t <f-exp>)
<f-exp-t> ::= #t
\end{lstlisting}

\subsection{Problem Description}

%
%

The problem definition includes as additional element the initial task network (line~\ref{l:tnihtn}) as in HDDL 1.0. But now, we propose to define in the initial state of the problem initial function values and initial time literals. It is also possible to define metric specs on solution plans. We have chosen here not to allow the definition of preferences. But preferences could be added in a next extension of the language. Likewise we have also chosen not to allow the definition of constraints on plans although it is possible in PDDL. We believe that it is preferable to add constraints on plans by adding logical constraints associated with task networks. This allows for a more unified language.

\begin{lstlisting}[firstnumber=last, escapechar=~]
<problem> ::= (define (problem <name>)
    (:domain <name>)
    [<require-def>]
    [<object-declaration>]
    [<htn>]~\label{l:tnihtn}~
    <init>
    [<goal>])~\label{l:goal}~
    [<metric-spec>]~\specReq{:numeric-fluents}
\end{lstlisting}

\begin{lstlisting}[firstnumber=last, escapechar=~]
<object-declaration> ::= ~\linebreak~(:objects <typed list (name)>)
<init> ::= (:init <init-el>*)
<init-el> ::= <literal (name)>
<init-el> ::=~\specReq{:timed-initial-literals}~(at <number> ~\mbox{<literal (name)>)}~
<init-el> ::=~\specReq{:numeric-fluents}~ (= ~\mbox{<basic-function-term>}~ <number>)
<basic-function-term> ::= <function-symbol>
<basic-function-term> ::= (<function-symbol> <name>*)
<goal> ::= (:goal <gd>)
\end{lstlisting}

The initial task network contains the definition of the problem class.

\begin{lstlisting}[firstnumber=last, escapechar=~]
<htn> ::= (:htn 
    [:parameters (<typed list (variable)>)] ~\label{l:tniparams}~
    <tasknetwork-def>)~\label{l:tnitasks}~
\end{lstlisting} 

%
%

The optional metric spec can be defined using the same syntax as in PDDL 3.0.

\begin{lstlisting}[firstnumber=last, escapechar=~]
<metric-spec> ::=~\specReq{:numeric-fluents}~ 
    (:metric <optimization> <metric-f-exp>)
<optimization> ::= minimize
<optimization> ::= maximize
<metric-f-exp> ::= (<binary-op>
    <metric-f-exp> <metric-f-exp>)
<metric-f-exp> ::= (<multi-op> 
    <metric-f-exp> <metric-f-exp>+)
<metric-f-exp> ::= (- <metric-f-exp>)
<metric-f-exp> ::= <number>
<metric-f-exp> ::= ( <function-symbol> 
    <name>* )
<metric-f-exp> ::= <function-symbol>
<metric-f-exp> ::= total-time
\end{lstlisting}

\section{Temporal and Numeric HDDL Requirements}

The overall definition includes all the following requirement flags as in HDDL 1.0:
\begin{itemize}
 \item \verb+:hierarchy+ requires the applied system to support HTN planning, so this can be seen as the basic requirement for the language defined here.
 \item \verb+:method-preconditions+ requires the applied system to support method preconditions or condition when define in durative methods.
\end{itemize}
But now, the following requirement flags are also compatible with HDDL 2.1:
\begin{description}
 \item \verb+:durative-actions+ requires the applied system to support durative actions and temporal ordering constraints in method definitions.
 \item \verb+:duration-inequalities+ requires the applied system to support duration inequalities in durative actions declarations. Implies \verb+:durative-actions+.
 \item \verb+:timed-initial-literals+ requires the applied system to support initial state with literals that become true at a specified time point. Implies \verb+:durative-actions+.
 \item \verb+:numeric-fluent+ requires the applied system to support numeric fluents in preconditions and effects of actions and methods.
 \item \verb+:continuous-effects+ requires the applied system to support continuous action effects.
 \end{description}
 We add the following requirement flag:
\begin{description}
\item \verb+:method-constraints+ requires to specify constraints on method task decomposition. 
 \end{description}


\end{document}